\documentclass{article}
\pdfoutput=1

\PassOptionsToPackage{numbers, compress}{natbib}

\usepackage[final]{neurips_2018}

\usepackage[utf8]{inputenc} \usepackage[T1]{fontenc}    \usepackage{hyperref}       \usepackage{url}            \usepackage{booktabs}       \usepackage{amsfonts}       \usepackage{nicefrac}       \usepackage{microtype}      \usepackage{graphicx}
\usepackage{paralist}
\usepackage[inline]{enumitem}
\usepackage{listings}
\usepackage[title]{appendix}
\usepackage{subcaption}
\usepackage{multirow}
\usepackage{csquotes}
\usepackage[disable]{todonotes}
\usepackage{pgfplots}
\usepgfplotslibrary{dateplot}
\usepackage{tikz}
\usetikzlibrary{positioning, shapes, fit, arrows}
\usetikzlibrary{external}

\title{Transfer Learning from Speaker Verification to Multispeaker Text-To-Speech Synthesis}

\newcommand{\compactand}{\hskip 2.5ex plus 5ex}
\author{
  Ye Jia\thanks{Equal contribution.}\compactand
  Yu Zhang\footnotemark[1]\compactand
  Ron J. Weiss\footnotemark[1]\compactand
  Quan Wang\compactand
  Jonathan Shen\compactand
  Fei Ren\compactand
  \\\textbf{
  Zhifeng Chen\compactand
  Patrick Nguyen\compactand
  Ruoming Pang\compactand
  Ignacio Lopez Moreno\compactand
  Yonghui Wu
  } \\
  Google Inc. \\
  \texttt{\{jiaye,ngyuzh,ronw\}@google.com}
}
\begin{document}

\maketitle

\begin{abstract}
We describe a neural network-based system for text-to-speech (TTS) synthesis that is able to generate speech audio in the voice of different speakers, including those unseen during training.
Our system consists of three independently trained components:
\begin{inparaenum}[(1)]
\item a \emph{speaker encoder network}, trained on a speaker verification task using an independent dataset of noisy speech without transcripts from thousands of speakers, to generate a fixed-dimensional embedding vector from only seconds of reference speech from a target speaker;
\item a sequence-to-sequence \emph{synthesis network} based on Tacotron~2 that generates a mel spectrogram from
text, conditioned on the speaker embedding;
\item an auto-regressive WaveNet-based \emph{vocoder network} that converts the mel spectrogram into time domain waveform samples.
\end{inparaenum}
We demonstrate that the
proposed model
is able to transfer the knowledge of speaker variability learned by the discriminatively-trained speaker encoder to the multispeaker TTS task, and is able to synthesize natural speech from speakers unseen during training. We quantify the importance of training the speaker encoder
on a large and diverse speaker set in order to obtain the best generalization performance. Finally, we show that randomly sampled speaker embeddings can be used to synthesize speech in the voice of novel speakers dissimilar from those used in training, indicating that the model has learned a high quality speaker representation.
\end{abstract}

\iffalse
\begin{abstract}
We describe a neural network-based system for text-to-speech (TTS) synthesis that is able to generate speech audio in the voice of many different speakers, including those unseen during training.
Our \emph{synthesis network} extends the Tacotron~2 sequence-to-sequence model, which generates a mel spectrogram from text, to additionally condition its output on a speaker embedding.
This embedding is computed using a \emph{speaker encoder network} which encodes a speech spectrogram into a fixed dimensional embedding vector.  The speaker encoder is trained independently on a speaker verification task using a separate dataset of noisy speech from thousands of different speakers without transcripts.
During training, the synthesis network is not explicitly given speaker identity labels.
We demonstrate that the synthesis network is able to transfer the knowledge of speaker variability learned by the discriminatively-trained speaker encoder to the new task, and is able to synthesize natural speech from speakers that were not seen during training using only a few seconds of reference audio.
We demonstrate the importance of training on a large and diverse speaker set in order to obtain the best generalization performance.
Finally, we observe that randomly sampled speaker embeddings can be used to synthesize speech in the voice of novel speakers dissimilar from those used in training, indicating that the model has learned a high quality speaker representation. \end{abstract}
\fi

\section{Introduction}

The goal of this work is to build a TTS system which can generate natural speech for a variety of speakers in a data efficient manner. We specifically address a zero-shot learning setting, where a few seconds of untranscribed reference audio from a target speaker is used to synthesize new speech in that speaker's voice, without updating any model parameters.
Such systems have accessibility applications, such as restoring the ability to communicate naturally to users who have lost their voice and are therefore unable to provide many new training examples. They could also enable new applications, such as transferring a voice across languages for more natural speech-to-speech translation, or generating realistic speech from text in low resource settings.
However, it is also important to note the potential for misuse of this technology, for example impersonating someone's voice without their consent.  In order to address safety concerns consistent with principles such as \cite{google2018aiprinciples}, we  verify that voices generated by the proposed model can easily be distinguished from real voices.

Synthesizing natural speech requires training on a large number of high quality speech-transcript pairs, and supporting many speakers usually uses tens of minutes of training data per speaker \cite{arik2017deepvoice2}. Recording a large amount of high quality data for many speakers is impractical. Our approach is to decouple speaker modeling from speech synthesis by independently training a speaker-discriminative embedding network that captures the space of speaker characteristics
and training a high quality TTS model on a smaller dataset conditioned on the representation learned by the first network.
Decoupling the networks enables them to be trained on independent data, which reduces the need to obtain high quality multispeaker training data.
We train the speaker embedding network on a speaker verification task to determine if two different utterances were spoken by the same speaker. In contrast to the subsequent TTS model, this network is trained on untranscribed speech containing reverberation and background noise from a large number of speakers.  

We demonstrate that the speaker encoder and synthesis networks can be trained on unbalanced and disjoint sets of speakers and still generalize well.  We train the synthesis network on 1.2K speakers and show that training the encoder on a much larger set of 18K speakers improves adaptation quality, and further enables synthesis of completely novel speakers by sampling from the embedding prior.

There has been significant interest in end-to-end training of TTS models,
which are trained directly from text-audio pairs, without depending on hand crafted intermediate representations \cite{Sotelo2017Char2wavES,yx2017tacotron}.
Tacotron 2~\cite{shen2018natural}
used WaveNet~\cite{van2016wavenet} as a vocoder to invert spectrograms generated by an encoder-decoder architecture with attention \cite{bahdanau2014neural},
obtaining naturalness approaching that of human speech
by combining
Tacotron's~\cite{yx2017tacotron} prosody with WaveNet's audio quality. 
It only supported a single speaker.

Gibiansky et al.~\cite{arik2017deepvoice2} introduced a multispeaker variation of Tacotron which learned low-dimensional speaker embedding for each training speaker.
Deep Voice 3 \cite{ping2018deepvoice3} proposed a fully convolutional encoder-decoder architecture which scaled up to support over 2,400~speakers from LibriSpeech~\cite{panayotov2015librispeech}.

These systems learn a fixed set of speaker embeddings and therefore only support synthesis of voices seen during training.  In contrast, VoiceLoop~\cite{taigman2018voiceloop} proposed a novel architecture based on a fixed size memory buffer which can generate speech from voices unseen during training.  Obtaining good results required tens of minutes of enrollment speech and transcripts for a new speaker.

Recent extensions have enabled few-shot speaker adaptation where only a few seconds of speech per speaker (without transcripts) can be used to generate new speech in that speaker's voice. \cite{arik2018neural} extends Deep Voice 3, comparing a \emph{speaker adaptation} method similar to \cite{taigman2018voiceloop} where the
model
parameters (including speaker embedding) are fine-tuned on a small amount of adaptation data
to a \emph{speaker encoding} method which uses a neural network to predict
speaker embedding
directly from a spectrogram.
The latter approach is significantly more data efficient, obtaining higher naturalness using small amounts of adaptation data,  in as few as one or two utterances. It is also significantly more computationally efficient since it does not require hundreds of backpropagation iterations.

Nachmani et al.~\cite{nachmani2018fitting} similarly extended VoiceLoop to utilize a target speaker encoding network to predict a speaker embedding.  This network is trained jointly with the synthesis network using a contrastive triplet loss to ensure that embeddings predicted from utterances by the same speaker are closer than embeddings computed from different speakers.  In addition, a cycle-consistency loss is used to ensure that the synthesized speech encodes to a similar embedding as the adaptation utterance.

A similar spectrogram encoder network, trained without a triplet loss, was shown to work for transferring target prosody to synthesized speech~\cite{rj2018transfer}.
In this paper we demonstrate that training a similar encoder to discriminate between speakers leads to reliable transfer of speaker characteristics. 
Our work is most similar to the speaker encoding models in~\cite{arik2018neural,nachmani2018fitting}, except that we utilize a network independently-trained for a speaker verification task on a large dataset of untranscribed audio from tens of thousands of speakers, using a state-of-the-art generalized end-to-end loss~\cite{wan2018generalized}.
\cite{nachmani2018fitting} incorporated a similar speaker-discriminative representation into their model, however all components were trained jointly.  In contrast, we explore transfer learning from a pre-trained speaker verification model.

Doddipatla et al.\ \cite{doddipatla2017speaker} used a similar transfer learning configuration where a speaker embedding computed from a pre-trained speaker classifier was used to condition a TTS system.  In this paper we utilize an end-to-end synthesis network which does not rely on intermediate linguistic features, and a substantially different speaker embedding network which is not limited to a closed set of speakers.
Furthermore, we analyze how quality varies with the number of speakers in the training set, and find that zero-shot transfer requires training on  thousands of speakers, many more than were used in \cite{doddipatla2017speaker}.

\section{Multispeaker speech synthesis model}
\label{sec.model}

\begin{figure}[t]
  \centering
  \begin{tikzpicture}[auto, font=\small, node distance=0.7cm and 0.5cm, >=latex']
    \pgfdeclarelayer{back}
    \pgfsetlayers{back,main}

    \tikzstyle{block} = [draw, fill=blue!20, align=center, rectangle, minimum height=2em, minimum width=4.5em]
    \tikzstyle{speakerenc block} = [block, fill=green!20]
    \tikzstyle{vocoder block} = [block, fill=red!20]
    \tikzstyle{input} = [align=center, inner sep=0]
    \tikzstyle{output} = []

    \node [input, name=inputaudio, align=center, font=\scriptsize\sffamily] {speaker \\ reference \\ waveform};
    \node [speakerenc block, right=0.8cm of inputaudio, inner ysep=1ex] (speakerenc) {Speaker \\ Encoder};
    
    \node [input, name=inputtext, below= of inputaudio, align=center, font=\scriptsize\sffamily] {grapheme or \\ phoneme \\ sequence};
    \node [block] (ttsenc) at (inputtext -| speakerenc) {Encoder};
    \node [block, right=0.25cm of ttsenc, minimum width=0] (concat) {concat};
    \node [block, right=0.25cm of concat] (ttsatt) {Attention};
    \node [block, right=0.25cm of ttsatt] (ttsdec) {Decoder};
    \node [above=0.09cm of ttsenc, inner sep=0, align=right, font=\small] (synthlabel) {\hspace{-0.5ex}Synthesizer};
    \begin{pgfonlayer}{back}
      \node [fill=blue!10, fit=(synthlabel)(ttsenc)(ttsdec), inner xsep=0.08cm] {};
    \end{pgfonlayer}
    \node [vocoder block, right=0.7cm of ttsdec] (vocoder) {Vocoder};
    \node [output, right= of vocoder, font=\scriptsize\sffamily] (output) {waveform};

    \draw [->] (inputaudio) -- (speakerenc);
    \draw [->] (speakerenc) -| node[align=left, font=\scriptsize\sffamily, color=darkgray] {speaker \\ embedding} (concat);
    
    \draw [->] (inputtext) -- (ttsenc);
    \draw [->] (ttsenc) -- (concat);
    \draw [->] (concat) -- (ttsatt);
    \draw [->] (ttsatt) -- (ttsdec);
    \draw [->] (ttsdec) edge[bend right=45] (ttsatt);
    \draw [->] (ttsdec) -- node [align=left, font=\scriptsize\sffamily, pos=1.2, color=darkgray] {log-mel \\ spectrogram \\ \vspace{1ex}} (vocoder);
    \draw [->] (vocoder) -- (output);
  \end{tikzpicture}
  \caption{Model overview.  Each of the three components are trained independently.}
  \label{fig.model_overview}
\end{figure}
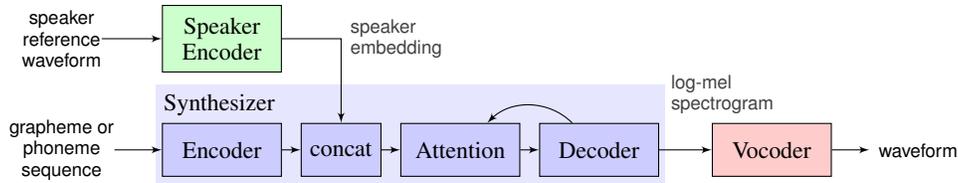

Our system is composed of three independently trained neural networks, illustrated in Figure~\ref{fig.model_overview}:
\begin{inparaenum}[(1)]
\item a recurrent \emph{speaker encoder}, based on \cite{wan2018generalized}, which computes a fixed dimensional vector from a speech signal, \item a sequence-to-sequence \emph{synthesizer}, based on \cite{shen2018natural}, which predicts a mel spectrogram from a sequence of grapheme
or phoneme inputs, conditioned on the speaker embedding vector,
and
\item an autoregressive WaveNet \cite{van2016wavenet}
\emph{vocoder}, which converts the spectrogram into time domain
waveforms.\footnote[1]{See
\url{https://google.github.io/tacotron/publications/speaker_adaptation} for samples.}

\end{inparaenum}

\subsection{Speaker encoder}
\label{sec.model.speaker_encoder}

The speaker encoder is used to condition the synthesis network on a reference speech signal from the desired target speaker.  Critical to good generalization is the use of a representation which captures the characteristics of different speakers, and the ability to identify these characteristics using only a short adaptation signal, independent of its phonetic content and background noise. These requirements are satisfied using a speaker-discriminative model trained on a text-independent speaker verification task.

We follow~\cite{wan2018generalized}, which proposed a highly scalable and accurate neural network framework for speaker verification. 
The network maps a sequence of log-mel spectrogram frames computed from a speech utterance of arbitrary length, to a fixed-dimensional embedding vector, known as \textit{d-vector} \cite{variani2014deep, heigold2016end}.
The network is trained to optimize a generalized end-to-end speaker verification
loss, so that embeddings of utterances from the same speaker have high cosine similarity, while those of utterances from different speakers are far apart in the embedding space. The training dataset consists of speech audio examples segmented into 1.6 seconds and associated speaker identity labels; no transcripts are used.

Input 40-channel log-mel spectrograms are passed to a network consisting of a stack of 3~LSTM layers of 768~cells, each followed by a projection to 256~dimensions.  The final embedding is created by $L_2$-normalizing the output of the top layer at the final frame.
During inference, an arbitrary length utterance is broken into 800ms windows, overlapped by 50\%.  The network is run independently on each window, and the outputs are averaged and normalized to create the final utterance embedding.

Although the network is not optimized directly to learn a representation which captures speaker characteristics relevant to synthesis, we find that training on a speaker discrimination task leads to an embedding which is directly suitable for conditioning the synthesis network on speaker identity.

\subsection{Synthesizer} \label{sec.model.synthesis}

We extend the recurrent sequence-to-sequence with attention Tacotron~2 architecture \cite{shen2018natural} to support  multiple speakers following a scheme similar to \cite{arik2017deepvoice2}.
An embedding vector for the target speaker is concatenated with the synthesizer encoder output at each time step.
In contrast to~\cite{arik2017deepvoice2}, 
we find that simply passing
embeddings to the attention layer, as in Figure~\ref{fig.model_overview}, converges across different speakers. 

We compare two variants of this model, one which computes the embedding using the speaker encoder, and a baseline which optimizes a fixed embedding for each speaker in the training set, essentially learning a lookup table of speaker embeddings similar to~\cite{arik2017deepvoice2,ping2018deepvoice3}.

The synthesizer is trained on pairs of text transcript and target audio.
At the input, we 
map the text to a sequence of phonemes, which leads to faster convergence and improved pronunciation of rare words and proper nouns. The network is trained in a transfer learning configuration, using a pretrained speaker encoder (whose parameters are frozen) to extract a speaker embedding from the target audio, i.e.\ the speaker reference signal is the same as the target speech during training.
No explicit speaker identifier labels are used during training.

Target spectrogram features are computed from 50ms windows computed with a 12.5ms step, passed through an 80-channel mel-scale filterbank followed by log dynamic range compression.
We extend \cite{shen2018natural} by augmenting the $L_2$ loss on the predicted spectrogram with an additional $L_1$ loss. In practice, we found this combined loss to be more robust on noisy training data.
In contrast to \cite{nachmani2018fitting}, we don't introduce additional loss terms based on the speaker embedding.

\subsection{Neural vocoder}
\label{sec.model.vocoder}

We use the sample-by-sample autoregressive WaveNet~\cite{van2016wavenet} as a vocoder to invert synthesized mel spectrograms emitted by the synthesis network into time-domain waveforms.  The architecture is the same as that described in~\cite{shen2018natural}, composed of 30 dilated convolution layers. The network is not directly conditioned on the output of the speaker encoder.  The mel spectrogram predicted by the synthesizer network captures all of the relevant detail needed for high quality synthesis of a variety of voices, allowing a multispeaker vocoder to be constructed by simply training on data from many speakers.

\subsection{Inference and zero-shot speaker adaptation}

During inference the model is conditioned using arbitrary untranscribed speech audio, which does not need to match the text to be synthesized.
Since the speaker characteristics to use for synthesis are inferred from audio, it can be conditioned on audio from speakers that are outside the training set.
In practice we find that using a single audio clip of a few seconds duration is sufficient to synthesize new speech with the corresponding speaker characteristics, representing zero-shot adaptation to novel speakers.
In Section~\ref{sec.experiments} we evaluate how well this process  generalizes to previously unseen speakers.

\begin{figure*}[t]
  \begin{center}
    \includegraphics[width=\textwidth]{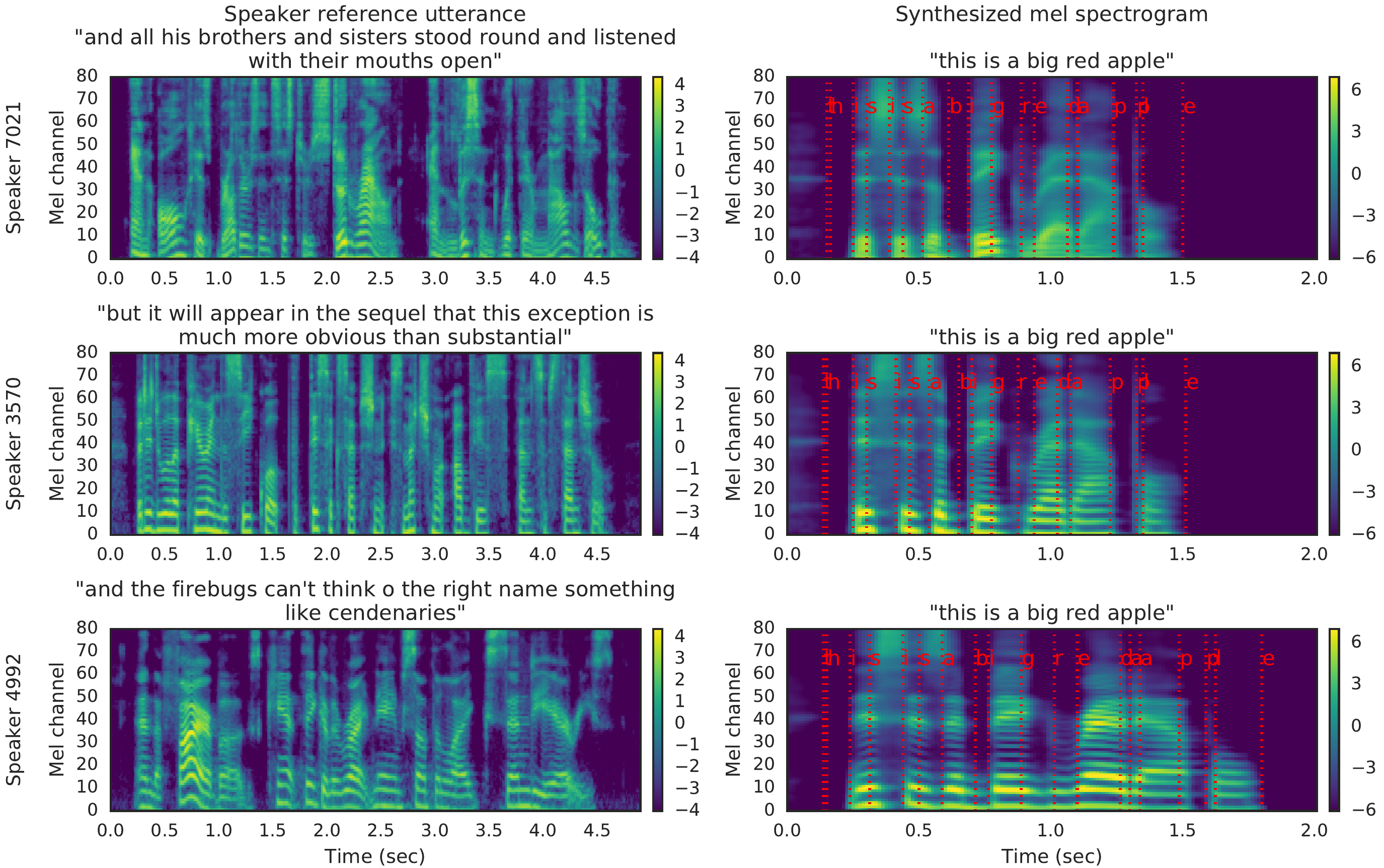}
    \caption{Example synthesis of a sentence in different voices using the proposed system.
    Mel spectrograms are visualized for 
    reference utterances used to generate speaker embeddings (left), and the corresponding
    synthesizer     outputs (right).
    The text-to-spectrogram alignment
    is shown in red.  Three speakers held out of the
    train sets are used: one male (top) and two female (center and bottom).
    }
  \label{fig.example_outputs}
  \end{center}
  \vskip-1.2ex
\end{figure*}

An example of the inference process is visualized in Figure~\ref{fig.example_outputs}, which shows spectrograms synthesized using several different 5 second speaker reference utterances.  Compared to those of the female (center and bottom) speakers, the synthesized male (top) speaker spectrogram has noticeably lower fundamental frequency, visible in the denser harmonic spacing (horizontal stripes) in low frequencies, as well as formants, visible in the mid-frequency peaks present during vowel sounds such as the `i' at 0.3 seconds -- the top male $F_2$ is in mel channel 35, whereas the $F_2$ of the middle speaker appears closer to channel 40.
Similar differences are also visible in sibilant sounds, e.g.\ the `s' at 0.4 seconds contains more energy in lower frequencies in the male voice than in the female voices.
Finally, the characteristic speaking rate is also captured to some extent by the speaker embedding, as can be seen by the longer signal duration in the bottom row compared to the top two.
Similar observations can be made about the corresponding reference utterance spectrograms in the right column. 
\section{Experiments}
\label{sec.experiments}

We used two public datasets for training the speech synthesis and vocoder networks.
    VCTK~\cite{veaux2017cstr} contains 44~hours of clean speech from 109 speakers, the majority of which have British accents. We downsampled the audio to 24~kHz, trimmed leading and trailing silence (reducing the median duration from 3.3 seconds to 1.8 seconds), and split into three subsets: train, validation (containing the same speakers as the train set) and test (containing 11 speakers held out from the train and validation sets).
    
    LibriSpeech~\cite{panayotov2015librispeech} consists of the union of the two ``clean'' training sets, comprising     436 hours of speech from 1,172 speakers, sampled at 16~kHz.  The majority of speech is US English, however since it is sourced from audio books, the tone and style of speech can differ significantly between utterances from the same speaker.
    We resegmented the data into shorter utterances by force aligning the audio to the transcript using an ASR model and breaking segments on silence, reducing the median duration from 14 to 5 seconds.     As in the original dataset, there is no punctuation in transcripts.
    The speaker sets are completely disjoint among the train, validation, and test sets.

Many recordings in the LibriSpeech clean corpus
contain noticeable environmental and stationary background noise. We preprocessed the target spectrogram using a simple spectral subtraction~\cite{boll1979suppression} denoising procedure, where the background noise spectrum of an utterance was estimated as the 10th percentile of the energy in each frequency band across the full signal.
This process was only used on the synthesis target; the original noisy speech was passed to the speaker encoder.

We trained separate synthesis and vocoder networks for each of these two corpora.
Throughout this section, we used synthesis networks trained on phoneme inputs, in order to control for pronunciation in subjective evaluations.
For the VCTK dataset, whose audio is quite clean, we found that the vocoder trained on ground truth mel spectrograms worked well. However for LibriSpeech, which is noisier, we found it necessary to train the vocoder on spectrograms predicted by the synthesizer network.  No denoising was performed on the target waveform for vocoder training.

The speaker encoder was trained on
a proprietary voice search corpus
containing 36M utterances with median duration of 3.9 seconds from 18K English speakers in the United States. This dataset is not transcribed, but contains anonymized speaker identities. 
It is never used to train synthesis networks.

We primarily rely on crowdsourced Mean Opinion Score (MOS) evaluations based on subjective listening tests. All our MOS evaluations are aligned to the \textit{Absolute Category Rating} scale~\cite{rec1996p}, with rating scores from 1 to 5 in 0.5~point increments. We use this framework to evaluate synthesized speech along two dimensions: its naturalness and similarity
to real speech from the target speaker.

\subsection{Speech naturalness}

We compared the naturalness of synthesized speech using synthesizers and vocoders trained on VCTK and LibriSpeech.  We constructed an evaluation set of 
100 phrases which do not appear in any training sets,
and evaluated two sets of speakers for each model: one composed of speakers included in the train set (Seen), and another composed of those that were held out (Unseen). We used 11 seen and unseen speakers for VCTK and 10 seen and unseen speakers for LibriSpeech (Appendix \ref{appx:eval_speaker_sets}).
For each speaker, we randomly chose one utterance with duration of about 5 seconds to use to compute the speaker embedding (see Appendix \ref{appx:speech_duration}).  Each phrase was synthesized for each speaker, for a total of about 1,000 synthesized utterances per evaluation.
Each sample was rated by a single rater, and each evaluation was conducted independently: the outputs of
different models were not compared directly.

\begin{table}[t!]
\caption{Speech naturalness Mean Opinion Score (MOS) with 95\% confidence intervals.}
\label{tbl:MOS-naturalness}
\begin{center}
\begin{small}
\begin{tabular}{ccccc}
\toprule
System & VCTK  Seen & VCTK Unseen & LibriSpeech  Seen & LibriSpeech Unseen\\
\midrule
Ground truth & 4.43 $\pm$ 0.05 & 4.49 $\pm$ 0.05 & 4.49 $\pm$ 0.05  & 4.42 $\pm$ 0.07 \\ 
Embedding table & 4.12 $\pm$ 0.06 & N/A & $3.90 \pm 0.06$ & N/A \\
Proposed model & 4.07 $\pm$ 0.06 & 4.20 $\pm$ 0.06  & $3.89 \pm 0.06$  & $4.12 \pm 0.05$ \\
\bottomrule
\end{tabular}

\end{small}
\end{center}
\vskip-1ex
\end{table}

Results are shown in Table~\ref{tbl:MOS-naturalness}, comparing the proposed model to baseline multispeaker models that utilize a lookup table of speaker embeddings similar to \cite{arik2017deepvoice2,ping2018deepvoice3}, but otherwise have identical architectures to the proposed synthesizer network.
The proposed model achieved about 4.0 MOS in all datasets,
with the VCTK model obtaining a MOS about 0.2 points higher than the LibriSpeech model when evaluated on seen speakers.
This is the consequence of two drawbacks of the LibriSpeech dataset:
\begin{inparaenum}[(i)]
\item the lack of punctuation in transcripts, which makes it difficult for the model to learn to pause naturally, and
\item the higher level of background noise compared to VCTK, some of which the synthesizer has learned to reproduce, despite denoising the training targets as described above. \end{inparaenum}

Most importantly, the audio generated by our model for unseen speakers is deemed to be at least as natural as that generated for seen speakers.  Surprisingly, the MOS on unseen speakers is higher than that of seen speakers, by as much as 0.2 points on LibriSpeech.
This is a consequence of the randomly selected reference utterance for each speaker, which sometimes contains uneven and non-neutral prosody.  In informal listening tests we found that the  prosody of the synthesized speech sometimes mimics that of the reference, similar to \cite{rj2018transfer}.
This effect is larger on LibriSpeech, which contains more varied prosody. This suggests that additional care must be taken to disentangle speaker identity from prosody within the synthesis network, perhaps by integrating a prosody encoder as in \cite{rj2018transfer,wang2018style}, or by training on randomly paired reference and target utterances from the same speaker.

\subsection{Speaker similarity}
\label{sec.experiments.speaker_similarity}

To evaluate how well the synthesized speech matches that from the target speaker,
we paired each synthesized utterance with
a randomly selected ground truth utterance from the same speaker.
Each pair is rated by one rater with the following instructions:
``You should not judge the content, grammar, or audio quality of the sentences; instead, just focus on the similarity of the speakers to one another.''

\begin{table}[t]
\vskip-1ex
\caption{Speaker similarity Mean Opinion Score (MOS) with 95\% confidence intervals.}
\label{tbl:speaker-similarity}
\begin{center}
\begin{small}
\begin{tabular}{cccc}
\toprule
System & Speaker Set & VCTK & LibriSpeech \\ \midrule
Ground truth & Same speaker & $4.67 \pm 0.04$ & $4.33 \pm 0.08$ \\ Ground truth &                Same gender & $2.25 \pm 0.07$ & $1.83 \pm 0.07$ \\ Ground truth &                Different gender & $1.15 \pm 0.04$ & $1.04 \pm 0.03$ \\ \midrule
Embedding table &  Seen  & 4.17 $\pm$ 0.06 & $3.70 \pm 0.08$ \\ Proposed model & Seen & 4.22 $\pm$ 0.06 & $3.28 \pm 0.08$ \\ \midrule
Proposed model & Unseen & $3.28 \pm 0.07$  & $3.03 \pm 0.09$ \\ \bottomrule
\end{tabular}
\end{small}
\end{center}
\vskip-2ex
\end{table}

Results are shown in Table~\ref{tbl:speaker-similarity}.
The scores for the VCTK model tend to be higher than those for LibriSpeech, reflecting the cleaner nature of the dataset. 
This is also evident in the higher ground truth baselines on VCTK.
For seen speakers on VCTK,
the proposed model performs about as well as the baseline which uses an embedding lookup table for speaker conditioning.
However, on LibriSpeech, the proposed model obtained a lower similarity MOS than the baseline,
which is likely due to the wider degree of within-speaker variation (Appendix \ref{appx:speaker_variation}), and background noise level in the dataset.

On unseen speakers, the proposed model obtains lower similarity between ground truth and synthesized speech. On VCTK, the similarity score of
3.28 is between ``moderately similar'' and ``very similar'' on the evaluation scale.
Informally, it is clear that the proposed model is able to transfer the broad strokes of the speaker characteristics for unseen speakers, clearly reflecting the correct gender, pitch, and formant ranges (as also visualized in Figure~\ref{fig.example_outputs}).  But the significantly reduced similarity scores on unseen speakers suggests that some nuances, e.g.\ related to characteristic prosody, are lost.

The speaker encoder is trained only on North American accented speech.
As a result, accent mismatch
constrains our performance on speaker similarity on VCTK since the rater instructions did not specify how to judge accents, so raters may consider a pair
to be from different speakers if the accents do not match.
Indeed, examination of rater comments shows that our model sometimes produced a different accent than the ground truth, which led to lower scores. However, a few raters commented that the tone and inflection of the voices sounded very similar despite differences in accent.

\begin{table}[t]
\vskip-2ex
\caption{Cross-dataset evaluation on naturalness and speaker similarity for unseen speakers.}
\label{tbl:cross_sim}
\begin{center}
\begin{small}
\begin{tabular}{cccc}
\toprule
Synthesizer Training Set & Testing Set & Naturalness & Similarity \\
\midrule
VCTK & LibriSpeech & $4.28 \pm 0.05$ & $1.82 \pm 0.08$ \\
LibriSpeech & VCTK & $4.01 \pm 0.06$ & $2.77 \pm 0.08$ \\
\bottomrule
\end{tabular}
\end{small}
\end{center}
\vskip-1ex
\end{table}

As an initial evaluation of the ability to generalize to out of domain speakers, we used synthesizers trained on VCTK and LibriSpeech to synthesize speakers from the other dataset. We only varied the train set of the synthesizer and vocoder networks; both models used an identical speaker encoder.
As shown in Table~\ref{tbl:cross_sim},
the models were able to generate speech with the same degree of naturalness as on unseen, but in-domain, speakers shown in Table~\ref{tbl:MOS-naturalness}.
However, the LibriSpeech model synthesized VCTK speakers with significantly higher speaker similarity  than the VCTK model is able to synthesize LibriSpeech speakers. The better generalization of the LibriSpeech model suggests that training the synthesizer on only 100 speakers is insufficient to enable high quality speaker transfer.

\subsection{Speaker verification}
\label{sec.experiments.sv}

\begin{table}
\caption{Speaker verification EERs of different synthesizers on unseen speakers. }
\label{tbl:EER}
\begin{center}
\begin{small}
\begin{tabular}{cccc}
\toprule
Synthesizer Training Set& Training Speakers  & SV-EER on VCTK & SV-EER on LibriSpeech \\
\midrule
 Ground truth
& -- & 1.53\% & 0.93\% \\
VCTK & 98 &   10.46\% & 29.19\% \\
LibriSpeech & 1.2K & 6.26\% & 5.08\% \\
\bottomrule
\end{tabular}
\end{small}
\end{center}
\end{table}

As an objective metric of the degree of speaker similarity between synthesized and ground truth audio for unseen speakers, we evaluated the ability of a limited speaker verification system to distinguish synthetic from real speech.
We trained a new \textit{eval-only} speaker encoder with the same network topology as Section~\ref{sec.model.speaker_encoder}, but using a different training set of 28M utterances from 113K speakers. Using a different model for evaluation ensured that metrics were not only valid on a specific speaker embedding space.
We enroll the voices of 21 real speakers: 11 speakers from VCTK, and 10 from LibriSpeech,  and score synthesized waveforms against the set of enrolled speakers. All enrollment and verification speakers are unseen during synthesizer training.
Speaker verification equal error rates (SV-EERs) are 
estimated by pairing each test utterance with each enrollment speaker. We synthesized 100 test utterances for each speaker, so 21,000 or 23,100 trials
were performed for each evaluation. 

As shown in Table~\ref{tbl:EER},
as long as the synthesizer was trained on a sufficiently large set of speakers, i.e.\ on LibriSpeech, the synthesized speech is typically most similar to the ground truth voices.
The LibriSpeech synthesizer obtains similar EERs of 5-6\% using reference speakers from both datasets, whereas the one trained on VCTK performs much worse, especially on out-of-domain LibriSpeech speakers.  These results are consistent with the subjective evaluation in Table~\ref{tbl:cross_sim}.

To measure the difficulty of discriminating between real and synthetic speech for the same speaker, we performed an additional evaluation with an expanded set of enrolled speakers including 10 synthetic versions of the 10 real LibriSpeech speakers.
On this 20 voice discrimination task we
obtain an EER of 2.86\%, demonstrating that, while the synthetic speech tends to be close to the target speaker (cosine similarity > 0.6, and as in Table~\ref{tbl:EER}), it is nearly always even closer to other synthetic utterances for the same speaker (similarity > 0.7).
From this we can conclude that the proposed model can generate speech that resembles the target speaker, but not well enough to be confusable with a real speaker.

\subsection{Speaker embedding space}

\begin{figure}[t]
  \begin{center}
    \includegraphics[width=\textwidth]{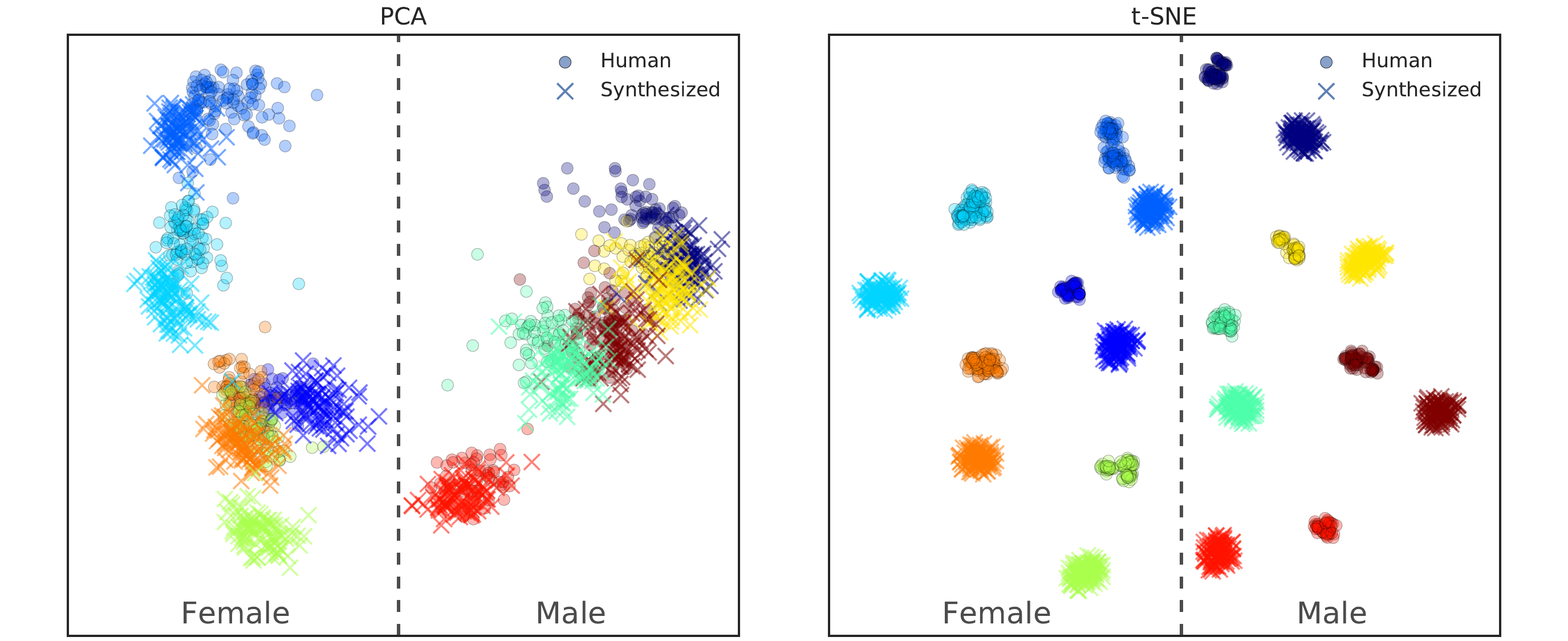}
    \caption{Visualization of speaker embeddings extracted from LibriSpeech utterances.  Each color corresponds to a different speaker. Real and synthetic utterances appear nearby when they are from the same speaker, however real and synthetic utterances consistently form distinct clusters.}
  \label{fig:librispeech_embedding}
  \end{center}
\end{figure}

Visualizing the speaker embedding space further contextualizes the quantitive results described in 
Section \ref{sec.experiments.speaker_similarity} and \ref{sec.experiments.sv}.
As shown in Figure \ref{fig:librispeech_embedding}, different speakers are well separated from each other in the speaker embedding space. The PCA visualization (left) shows that synthesized utterances tend to lie very close to real speech from the same speaker in the embedding space. However, synthetic utterances are still easily distinguishable from the real human speech as demonstrated by the t-SNE visualization (right) where utterances from each synthetic speaker form a distinct cluster adjacent to a cluster of real utterances from the corresponding speaker.

Speakers appear to be well separated by gender in both the PCA and t-SNE visualizations, with all female speakers appearing on the left, and all male speakers appearing on the right. This is an indication that the speaker encoder has learned a reasonable representation of speaker space.

\subsection{Number of speaker encoder training speakers}
\label{sec.experiments.num_speakers}

It is likely that the ability of the proposed model to generalize well across a wide variety of speakers is based on the quality of the representation learned by the speaker encoder.  We therefore explored the effect of the  speaker encoder training set on synthesis quality.  We made use of three additional training sets:
\begin{inparaenum}[(1)]
\item LibriSpeech Other, which contains 461 hours of speech from a set of 1,166 speakers disjoint from those in the clean subsets,
\item VoxCeleb~\cite{nagrani2017voxceleb}, and
\item VoxCeleb2~\cite{chung2018voxceleb2} which contain 139K utterances from 1,211 speakers, and 1.09M utterances from 5,994 speakers, respectively.
\end{inparaenum}

\begin{table}[t]
\vskip-2ex
\caption{Performance using speaker encoders (SEs) trained on different datasets.  Synthesizers are all trained on LibriSpeech Clean and
evaluated on held out speakers. LS: LibriSpeech, VC: VoxCeleb.}
\label{tbl:num_speakers}
\begin{center}
\begin{small}
\begin{tabular}{cccccc}
\toprule
SE Training Set & Speakers & Embedding Dim & Naturalness & Similarity & SV-EER \\ \midrule
LS-Clean &            1.2K & 64 & $3.73 \pm 0.06$ &
           $2.23 \pm 0.08$
           & 16.60\%\\
LS-Other &            1.2K & 64 & $3.60 \pm 0.06$ &
           $2.27 \pm 0.09$
           & 15.32\% \\
LS-Other + VC &            2.4K & 256 & $3.83 \pm 0.06$ &
           $2.43 \pm 0.09$
           & 11.95\% \\
LS-Other + VC + VC2 &            8.4K & 256 & $3.82 \pm 0.06$ &
           $2.54 \pm 0.09$ 
           & 10.14\% \\
Internal  & 18K & 256 & $4.12 \pm 0.05$ &
           $3.03 \pm 0.09$
           & 5.08\% \\
\bottomrule
\end{tabular}
\end{small} 
\end{center}
\end{table}

Table~\ref{tbl:num_speakers} compares the performance of the proposed model as a function of the number of speakers used to train the speaker encoder. This measures the importance of speaker diversity when training the speaker encoder.
To avoid overfitting, the speaker encoders trained on small datasets (top two rows) use a smaller network architecture (256-dim LSTM cells with 64-dim projections) and output 64 dimensional speaker embeddings.

We first evaluate the speaker encoder trained on LibriSpeech Clean and Other sets, each of which contain a similar number of speakers. In Clean, the speaker encoder and synthesizer are trained on the same data, a baseline similar to the non-fine tuned speaker encoder from \cite{arik2018neural}, except that it is trained discriminatively as in \cite{nachmani2018fitting}. 
This matched condition gives a slightly better naturalness and a similar similarity score. 
As the number of training speakers increases, both naturalness and similarity improve significantly. The objective EER results also improve alongside the subjective evaluations.

These results have an important implication for multispeaker TTS training. The data requirement for the speaker encoder is much cheaper than full TTS training since no transcripts are necessary, and the audio quality can be lower than for TTS training. We have shown that it is possible to synthesize very natural TTS by combining a speaker encoder network trained on large amounts of untranscribed data with a TTS network trained on a smaller set of high quality data.

\subsection{Fictitious speakers}

\begin{table}[t]
\caption{Speech from fictitious speakers compared to their nearest neighbors in the train sets. Synthesizer was trained on LS Clean. Speaker Encoder was trained on LS-Other + VC + VC2.} \label{tbl:fictitious_speakers}
\begin{center}
\begin{small}
\begin{tabular}{cccccc}
\toprule
Nearest neighbors in & Cosine similarity & SV-EER & Naturalness MOS \\
\midrule
Synthesizer train set & 0.222  &  56.77\% & \multirow{2}{*}{$3.65 \pm 0.06$}  \\
Speaker Encoder train set & 0.245  &  38.54\% \\
\bottomrule
\end{tabular}
\end{small}
\end{center}
\vskip-2ex
\end{table}

Bypassing the speaker encoder network and conditioning the synthesizer on random points in the speaker embedding space results in speech from fictitious speakers which are not present in
the train or test sets of either the synthesizer or the speaker encoder.
This is demonstrated in Table~\ref{tbl:fictitious_speakers}, which compares 10 such speakers, generated from uniformly sampled points on the surface of the unit hypersphere, to their nearest neighbors in the training sets of the component networks.
SV-EERs are computed using the same setup as Section~\ref{sec.experiments.sv} after enrolling voices of the 10 nearest neighbors.
Even though these speakers are totally fictitious, the synthesizer and the vocoder are able to generate audio as natural as for seen or unseen real speakers. The low cosine similarity  to the nearest neighbor training utterances and very high EER indicate that they are indeed distinct from the 
training speakers.

\section{Conclusion}

We present a neural network-based system for multispeaker TTS synthesis. The system combines an independently trained speaker encoder network with a sequence-to-sequence TTS synthesis network and neural vocoder based on Tacotron~2.
By leveraging the knowledge learned by the discriminative speaker encoder, the synthesizer is able to generate high quality speech not only for speakers seen during training, but also for speakers never seen before.
Through evaluations based on a speaker verification system as well as subjective listening tests, we demonstrated that the synthesized speech is reasonably similar to real speech from the target speakers, even on such unseen speakers.

We ran experiments to analyze the impact of the amount of data used to train the different components,
and found that, given sufficient speaker diversity in the synthesizer training set, speaker transfer quality could be significantly improved by increasing the amount of speaker encoder training data.

Transfer learning is critical to achieving these results.  By separating the training of the speaker encoder and the synthesizer, the system significantly lowers the requirements for multispeaker TTS training data. It requires neither speaker identity labels for the synthesizer training data, nor high quality clean speech or transcripts for the speaker encoder training data.  In addition, training the components independently significantly simplifies the training configuration of the synthesizer network compared to \cite{nachmani2018fitting} since it does not require additional triplet or contrastive losses. However, modeling speaker variation using a low dimensional vector limits the ability to leverage large amounts of reference speech.  Improving speaker similarity given more than a few seconds of reference speech requires a model adaptation approach as in \cite{arik2018neural}, and more recently in \cite{chen2018sample}.

Finally, we demonstrate that the model is able to generate realistic speech from fictitious speakers that are dissimilar from the training set, implying that the model has learned to utilize a realistic representation of the space of speaker variation.

The proposed model does not attain human-level naturalness, despite the use of a WaveNet vocoder (along with its very high inference cost), in contrast to the single speaker results from \cite{shen2018natural}.  This is a consequence of the additional difficulty of generating speech for a variety of speakers given significantly less data per speaker, as well as the use of datasets with lower data quality.
An additional limitation lies in the model's inability to transfer accents.  Given sufficient training data, this could be addressed by conditioning the synthesizer on independent speaker and accent embeddings.  
Finally, we note that the model is also not able to completely isolate the speaker voice from the prosody of the reference audio, a similar trend to that observed in \cite{rj2018transfer}.

\subsubsection*{Acknowledgements}

The authors thank Heiga Zen, Yuxuan Wang, Samy Bengio, the Google AI Perception team, and the Google TTS and DeepMind Research teams for their helpful discussions and feedback.

\bibliographystyle{plain}
\bibliography{references}

\begin{thebibliography}{10}

\bibitem{google2018aiprinciples}
{Artificial Intelligence at Google -- Our Principles}.
\newblock \url{https://ai.google/principles/}, 2018.

\bibitem{arik2018neural}
Sercan~O Arik, Jitong Chen, Kainan Peng, Wei Ping, and Yanqi Zhou.
\newblock Neural voice cloning with a few samples.
\newblock {\em arXiv preprint arXiv:1802.06006}, 2018.

\bibitem{bahdanau2014neural}
Dzmitry Bahdanau, Kyunghyun Cho, and Yoshua Bengio.
\newblock Neural machine translation by jointly learning to align and
  translate.
\newblock In {\em Proceedings of ICLR}, 2015.

\bibitem{boll1979suppression}
Steven Boll.
\newblock Suppression of acoustic noise in speech using spectral subtraction.
\newblock {\em IEEE Transactions on Acoustics, Speech, and Signal Processing},
  27(2):113--120, 1979.

\bibitem{chen2018sample}
Yutian Chen, Yannis Assael, Brendan Shillingford, David Budden, Scott Reed,
  Heiga Zen, Quan Wang, Luis~C Cobo, Andrew Trask, Ben Laurie, et~al.
\newblock Sample efficient adaptive text-to-speech.
\newblock {\em arXiv preprint arXiv:1809.10460}, 2018.

\bibitem{chung2018voxceleb2}
Joon~Son Chung, Arsha Nagrani, and Andrew Zisserman.
\newblock {VoxCeleb2}: Deep speaker recognition.
\newblock In {\em Interspeech}, pages 1086--1090, 2018.

\bibitem{doddipatla2017speaker}
Rama Doddipatla, Norbert Braunschweiler, and Ranniery Maia.
\newblock Speaker adaptation in dnn-based speech synthesis using d-vectors.
\newblock In {\em Proc. Interspeech}, pages 3404--3408, 2017.

\bibitem{arik2017deepvoice2}
Andrew Gibiansky, Sercan Arik, Gregory Diamos, John Miller, Kainan Peng, Wei
  Ping, Jonathan Raiman, and Yanqi Zhou.
\newblock {Deep Voice 2}: Multi-speaker neural text-to-speech.
\newblock In I.~Guyon, U.~V. Luxburg, S.~Bengio, H.~Wallach, R.~Fergus,
  S.~Vishwanathan, and R.~Garnett, editors, {\em Advances in Neural Information
  Processing Systems 30}, pages 2962--2970. Curran Associates, Inc., 2017.

\bibitem{heigold2016end}
Georg Heigold, Ignacio Moreno, Samy Bengio, and Noam Shazeer.
\newblock End-to-end text-dependent speaker verification.
\newblock In {\em Acoustics, Speech and Signal Processing (ICASSP), 2016 IEEE
  International Conference on}, pages 5115--5119. IEEE, 2016.

\bibitem{nachmani2018fitting}
Eliya Nachmani, Adam Polyak, Yaniv Taigman, and Lior Wolf.
\newblock Fitting new speakers based on a short untranscribed sample.
\newblock {\em arXiv preprint arXiv:1802.06984}, 2018.

\bibitem{nagrani2017voxceleb}
Arsha Nagrani, Joon~Son Chung, and Andrew Zisserman.
\newblock {VoxCeleb}: A large-scale speaker identification dataset.
\newblock {\em arXiv preprint arXiv:1706.08612}, 2017.

\bibitem{panayotov2015librispeech}
Vassil Panayotov, Guoguo Chen, Daniel Povey, and Sanjeev Khudanpur.
\newblock {LibriSpeech}: an {ASR} corpus based on public domain audio books.
\newblock In {\em Acoustics, Speech and Signal Processing (ICASSP), 2015 IEEE
  International Conference on}, pages 5206--5210. IEEE, 2015.

\bibitem{ping2018deepvoice3}
Wei Ping, Kainan Peng, Andrew Gibiansky, Sercan~O. Arik, Ajay Kannan, Sharan
  Narang, Jonathan Raiman, and John Miller.
\newblock {Deep Voice 3}: 2000-speaker neural text-to-speech.
\newblock In {\em Proc. International Conference on Learning Representations
  ({ICLR})}, 2018.

\bibitem{rec1996p}
ITUT Rec.
\newblock P. 800: Methods for subjective determination of transmission quality.
\newblock {\em International Telecommunication Union, Geneva}, 1996.

\bibitem{shen2018natural}
Jonathan Shen, Ruoming Pang, Ron~J. Weiss, Mike Schuster, Navdeep Jaitly,
  Zongheng Yang, Zhifeng Chen, Yu~Zhang, Yuxuan Wang, RJ~Skerry-Ryan, Rif~A.
  Saurous, Yannis Agiomyrgiannakis, and Yonghui. Wu.
\newblock Natural {TTS} synthesis by conditioning {WaveNet} on mel spectrogram
  predictions.
\newblock In {\em Proc. {IEEE} International Conference on Acoustics, Speech,
  and Signal Processing ({ICASSP})}, 2018.

\bibitem{rj2018transfer}
RJ~Skerry-Ryan, Eric Battenberg, Ying Xiao, Yuxuan Wang, Daisy Stanton, Joel
  Shor, Ron~J. Weiss, Rob Clark, and Rif~A. Saurous.
\newblock Towards end-to-end prosody transfer for expressive speech synthesis
  with {T}acotron.
\newblock {\em arXiv preprint arXiv:1803.09047}, 2018.

\bibitem{Sotelo2017Char2wavES}
Jose Sotelo, Soroush Mehri, Kundan Kumar, Jo{\~a}o~Felipe Santos, Kyle Kastner,
  Aaron Courville, and Yoshua Bengio.
\newblock {Char2Wav}: End-to-end speech synthesis.
\newblock In {\em Proc. International Conference on Learning Representations
  ({ICLR})}, 2017.

\bibitem{taigman2018voiceloop}
Yaniv Taigman, Lior Wolf, Adam Polyak, and Eliya Nachmani.
\newblock {VoiceLoop}: Voice fitting and synthesis via a phonological loop.
\newblock In {\em Proc. International Conference on Learning Representations
  ({ICLR})}, 2018.

\bibitem{van2016wavenet}
A{\"a}ron van~den Oord, Sander Dieleman, Heiga Zen, Karen Simonyan, Oriol
  Vinyals, Alex Graves, Nal Kalchbrenner, Andrew Senior, and Koray Kavukcuoglu.
\newblock {WaveNet}: A generative model for raw audio.
\newblock {\em CoRR abs/1609.03499}, 2016.

\bibitem{variani2014deep}
Ehsan Variani, Xin Lei, Erik McDermott, Ignacio~Lopez Moreno, and Javier
  Gonzalez-Dominguez.
\newblock Deep neural networks for small footprint text-dependent speaker
  verification.
\newblock In {\em Acoustics, Speech and Signal Processing (ICASSP), 2014 IEEE
  International Conference on}, pages 4052--4056. IEEE, 2014.

\bibitem{veaux2017cstr}
Christophe Veaux, Junichi Yamagishi, Kirsten MacDonald, et~al.
\newblock {CSTR VCTK Corpus}: English multi-speaker corpus for {CSTR} voice
  cloning toolkit, 2017.

\bibitem{wan2018generalized}
Li~Wan, Quan Wang, Alan Papir, and Ignacio~Lopez Moreno.
\newblock Generalized end-to-end loss for speaker verification.
\newblock In {\em Proc. {IEEE} International Conference on Acoustics, Speech,
  and Signal Processing ({ICASSP})}, 2018.

\bibitem{yx2017tacotron}
Yuxuan Wang, RJ~Skerry-Ryan, Daisy Stanton, Yonghui Wu, Ron~J. Weiss, Navdeep
  Jaitly, Zongheng Yang, Ying Xiao, Zhifeng Chen, Samy Bengio, Quoc Le, Yannis
  Agiomyrgiannakis, Rob Clark, and Rif~A. Saurous.
\newblock Tacotron: Towards end-to-end speech synthesis.
\newblock In {\em Proc. Interspeech}, pages 4006--4010, August 2017.

\bibitem{wang2018style}
Yuxuan Wang, Daisy Stanton, Yu~Zhang, RJ~Skerry-Ryan, Eric Battenberg, Joel
  Shor, Ying Xiao, Fei Ren, Ye~Jia, and Rif~A Saurous.
\newblock Style tokens: Unsupervised style modeling, control and transfer in
  end-to-end speech synthesis.
\newblock {\em arXiv preprint arXiv:1803.09017}, 2018.

\end{thebibliography}

\newpage

\begin{appendices}

\section{Additional joint training baselines}

\begin{table}[h]
\caption{Speech naturalness and speaker similarity Mean Opinion Score (MOS) with 95\% confidence intervals of baseline models where the speaker encoder and synthesizer networks are trained jointly (top two rows).
Included for comparison are the separately trained baseline from Table~\ref{tbl:num_speakers} (middle row) as well as the embedding lookup table baseline and proposed model from Tables~\ref{tbl:MOS-naturalness} and \ref{tbl:speaker-similarity} (bottom two rows).
All but the bottom row, are trained entirely on LibriSpeech. The bottom row uses a speaker encoder trained on a separate speaker corpus. All evaluations are on LibriSpeech.}
\label{tbl:MOS-joint_training_baselines}
\begin{center}
\begin{small}
\begin{tabular}{lccccc}
\toprule  
& & \multicolumn{2}{c}{Naturalness MOS} & \multicolumn{2}{c}{Similarity MOS} \\ 
System & \hspace{-2em}Embedding Dim\hspace{-1em} & Seen & Unseen & Seen & Unseen\\
\midrule
Joint training & \hspace{-1em}64
& 3.72 $\pm$ 0.06 & 3.59 $\pm$ 0.07
  & 2.47 $\pm$ 0.08 & 2.44 $\pm$ 0.09 \\
Joint training + speaker loss & \hspace{-1em}64
& 3.71 $\pm$ 0.06 & 3.71 $\pm$ 0.06
  & 2.82 $\pm$ 0.08 & 2.12 $\pm$ 0.08 \\
Separate training (Table~\ref{tbl:num_speakers}) & \hspace{-1em}64
& 3.88 $\pm$ 0.06 & 3.73 $\pm$ 0.06
 & 2.64 $\pm$ 0.08 & 2.23 $\pm$ 0.08 \\
Embedding table (Tables~\ref{tbl:MOS-naturalness},\ref{tbl:speaker-similarity}) & \hspace{-1em}64
& $3.90 \pm 0.06$ & N/A 
  & $3.70 \pm 0.08$ & N/A \\
\midrule
Proposed model (Tables~\ref{tbl:MOS-naturalness},\ref{tbl:speaker-similarity},\ref{tbl:num_speakers}) & \hspace{-1em}256
& $3.89 \pm 0.06$ & $4.12 \pm 0.05$ 
  & $3.28 \pm 0.08$ & $3.03 \pm 0.09$ \\
\bottomrule
\end{tabular}
\end{small}
\end{center}
\end{table}

Although separate training of the speaker encoder and synthesizer networks is necessary if the speaker encoder is trained on a larger corpus of untranscribed speech, as described in Section~\ref{sec.experiments.num_speakers}, in this section we evaluate the effectiveness of joint training of the speaker encoder and synthesizer networks as a baseline, similar to \cite{nachmani2018fitting}.

We train on the Clean subset of LibriSpeech, containing 1.2K speakers, and use a speaker embedding dimension of 64 following Section~\ref{sec.experiments.num_speakers}. We compare two baseline jointly-trained systems: one without any constraints on the output of the speaker encoder, analogous to \cite{rj2018transfer}, and another with an additional speaker discrimination loss formed by passing the 64 dimension speaker embedding through a linear projection to form the logits for a softmax speaker classifier, optimizing a corresponding cross-entropy loss.

Naturalness and speaker similarity MOS results are shown in Table~\ref{tbl:MOS-joint_training_baselines}, comparing these jointly trained baselines to results reported in previous sections.
We find that both jointly trained models obtain similar naturalness MOS on Seen speakers, with the variant incorporating a discriminative speaker loss performing better on Unseen speakers.
In terms of both naturalness and similarity on Unseen speakers, the model which includes the speaker loss has nearly the same performance as the baseline from Table~\ref{tbl:num_speakers}, which uses a separately trained speaker encoder that is also optimized to discriminate between speakers.
Finally, we note that the proposed model, which uses a speaker encoder trained separately on a corpus of 18K speakers, significantly outperforms all baselines, once again highlighting the effectiveness of transfer learning for this task.

\section{Speaker variation}
\label{appx:speaker_variation}

\todo{jiaye: can we add confidence intervals here, post submission. It would be interesting to see if some speakers have much higher variance.}
\begin{table}[h]
\caption{Ground truth MOS evaluations breakdown on unseen speakers. Similarity evaluations compare two utterances by the same speaker.}
\label{tbl:similarity_variation}
\begin{center}
\begin{small}

\begin{subtable}[t]{0.49\textwidth}
\caption{VCTK}
\begin{tabular}{cccc}
\toprule
Speaker & Gender & Naturalness & Similarity \\
\midrule
p230 & F & 4.22 & 4.65 \\
p240 & F & 4.57 & 4.67 \\
p250 & F & 4.31 & 4.72 \\
p260 & M & 4.56 & 4.31 \\
p270 & M & 4.29 & 4.77 \\
p280 & F & 4.41 & 4.71 \\
p300 & F & 4.60 & 4.87 \\
p310 & F & 4.56 & 4.52 \\
p330 & F & 4.34 & 4.77 \\
p340 & F & 4.44 & 4.71 \\
p360 & M & 4.36 & 4.63 \\
\bottomrule
\end{tabular}
\end{subtable}
\hfill
\begin{subtable}[t]{0.49\textwidth}
\caption{LibriSpeech}
\begin{tabular}{cccc}
\toprule
Speaker & Gender & Naturalness & Similarity \\
\midrule
1320 & M & 4.64 & 4.43 \\
2300 & M & 4.67 & 4.22 \\
3570 & F & 4.31 & 4.38 \\
3575 & F & 4.59 & 4.36 \\
4970 & F & 3.77 & 4.16 \\
4992 & F & 4.40 & 3.81 \\
6829 & F & 4.24 & 4.39 \\
7021 & M & 4.71 & 4.55 \\
7729 & M & 4.55 & 4.48 \\
8230 & M & 4.65 & 4.70 \\
\bottomrule
\end{tabular}
\end{subtable}

\end{small}
\end{center}
\end{table}

The tone and style of LibriSpeech utterances varies significantly between utterances even from the same speaker. In some examples, the speaker even tries to mimic a voice in a different gender. As a result, comparing the speaker similarity between different utterances from a same speaker (i.e.\ self-similarity) can sometimes be relatively low, and varies significantly speaker by speaker. Because of the noise level in LibriSpeech recordings, some speakers have significantly lower naturalness scores. This again varies significantly speaker by speaker.  This can be seen in Table~\ref{tbl:similarity_variation}.  In contrast, VCTK is more consistent in terms of both naturalness and self-similarity.

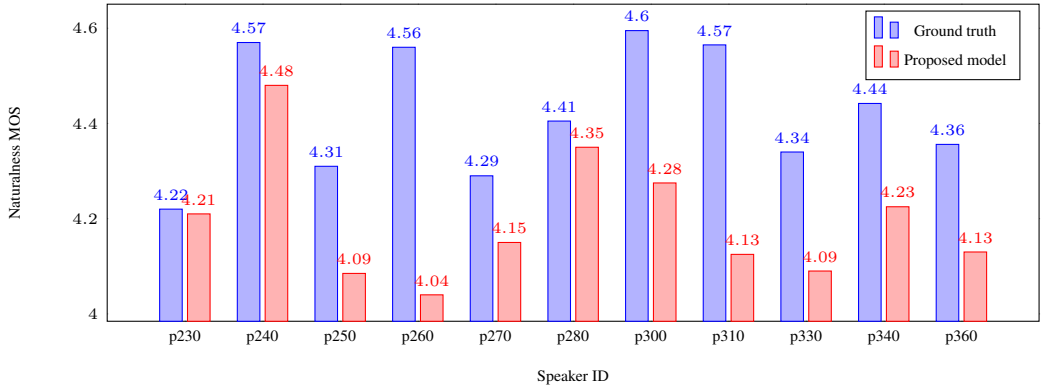
\begin{figure}[h]
    \centering
    \begin{tikzpicture}
      \begin{axis}[
        xlabel={\tiny Speaker ID},
        ylabel={\tiny Naturalness MOS},
        symbolic x coords = {p230, p240, p250, p260, p270, p280, p300, p310, p330, p340, p360},
        xtick = {p230, p240, p250, p260, p270, p280, p300, p310, p330, p340, p360},
        enlargelimits=true,
        x tick label style={font=\tiny, align=center},
        y tick label style={font=\tiny},
        nodes near coords,
        every node near coord/.append style={font=\tiny},
        width=\textwidth,
        height=5.8cm,
        bar width=0.3cm,
        tickwidth         = 0pt,
        enlargelimits=true,
        enlarge x limits  = 0.1,
        ybar]
        
      \addplot coordinates {(p230, 4.220) (p240, 4.570) (p250, 4.310) (p260, 4.560) (p270, 4.290) (p280, 4.405) (p300, 4.595) (p310, 4.565) (p330, 4.340) (p340, 4.442) (p360, 4.356)};
      \addplot coordinates {(p230, 4.210) (p240, 4.480) (p250, 4.085) (p260, 4.040) (p270, 4.150) (p280, 4.350) (p300, 4.275) (p310, 4.125) (p330, 4.090) (p340, 4.225) (p360, 4.130)}; 
      \legend{\tiny Ground truth, \tiny Proposed model}
      \end{axis}
    \end{tikzpicture}
    \caption{Per-speaker naturalness MOS of ground truth and synthesized speech on unseen VCTK speakers.}
    \label{tbl:spk-mos}
\end{figure}

Table~\ref{tbl:spk-mos} shows the variance in naturalness MOS across different speakers on synthesized audio.
It compares the MOS of different speakers for both ground truth and synthesized on VCTK, revealing that the performance of our proposed model on VCTK is also very speaker dependant. For example, speaker ``p240'' obtained a MOS of 4.48, which is very close to the MOS of the ground truth (4.57), but speaker ``p260'' is a full 0.5 points behind its ground truth.

\newpage

\section{Impact of reference speech duration}
\label{appx:speech_duration}

\begin{table}[h]
\caption{Impact of duration of reference speech utterance. Evaluated on VCTK.}
\label{tbl:source_len}
\begin{center}
\begin{small}
\begin{tabular}{cccccc}
\toprule
& 1 sec 
& 2 sec & 3 sec & 5 sec & 10 sec \\
\midrule
Naturalness (MOS) & $4.28 \pm 0.05$ & 
    $4.26 \pm 0.05$ & $4.18 \pm 0.06$ & $4.20 \pm 0.06$ & $4.16 \pm 0.06$ \\
Similarity (MOS) & $2.85 \pm 0.07$ & $3.17 \pm 0.07$ & $3.31 \pm 0.07$ & $3.28 \pm 0.07$ & $3.18 \pm 0.07$ \\
SV-EER & 17.28\% & 11.30\% & 10.80\% & 10.46\% & 11.50\% \\
\bottomrule
\end{tabular}
\end{small}
\end{center}
\vskip -0.1in
\end{table}

The proposed model depends on a reference speech signal fed into the speaker encoder. As shown in Table~\ref{tbl:source_len}, increasing the length of the reference speech significantly improved the similarity, because we can compute more precise speaker embedding with it. Quality saturates at about 5~seconds on VCTK.
Shorter reference utterances give slightly better naturalness, because they better match  the durations
of reference utterances used to train the synthesizer, whose median duration is 1.8 seconds. The proposed model achieves close to the best performance using only 2 seconds of reference audio.
The performance saturation using only 5~seconds of speech highlights a limitation of the proposed model, which is constrained by the small capacity of the speaker embedding. Similar scaling was found in \cite{arik2018neural}, where adapting a speaker embedding alone was shown to be effective given limited adaptation data, however fine tuning the full model was required to improve performance if more data was available. This pattern was also confirmed in more recent work \cite{chen2018sample}.

\iffalse
\section{Speaker embedding space}
\label{appx:speaker_emb}
\begin{figure}[h]
  \begin{center}
    \includegraphics[width=\textwidth]{figs/librispeech_embedding_space.pdf}
    \caption{Visualization of speaker embeddings extracted from LibriSpeech utterances.  Real (o) and synthetic (x) utterances appear nearby when they are from the same speaker, however real and synthetic utterances consistently form distinct clusters.}
  \label{fig:librispeech_embedding}
  \end{center}
\end{figure}

As visualized in Figure \ref{fig:librispeech_embedding}, different speakers are well separated from each other in the speaker embedding space. As shown in the PCA visualization (left), synthesized speech is very close to real speech from the same speaker in the embedding space. However they are still easily distinguishable from the real human speech as shown in the t-SNE visualization (right).

It's also interesting to note that in the PCA visualization\todo{It looks to me like you can see a similar separation in the t-SNE plot.  You just have to draw the corresponding line...} in Figure \ref{fig:librispeech_embedding}, speakers are separated by gender, with all female speakers appearing on the left, and all male speakers appearing on the right. This is an indication that the speaker encoder has learned a reasonable representation of speaker space.

\fi

\section{Evaluation speaker sets}
\label{appx:eval_speaker_sets}

\begin{table}[h!]
\caption{Speaker sets used for evaluation.}
\label{tbl:eval_speaker_sets}
\begin{center}
\begin{small}
\begin{subtable}[t]{\textwidth}
\begin{center}
\caption{VCTK}
\begin{tabular}{cccccccccccc}
\toprule
\multicolumn{12}{c}{Seen} \\
Speaker & p231 & p241 & p251 & p261 & p271 & p281 & p301 & p311 & p341 & p351 & p361 \\
Gender & F & M & M & F & M & M &F & M & F & F & F \\
\midrule
\multicolumn{12}{c}{Unseen} \\
Speaker & p230 & p240 & p250 & p260 & p270 & p280 & p300 & p310 & p330 & p340 & p360 \\
Gender & F & F & F & M & M & F & F & F & F & F & M \\
\bottomrule
\end{tabular}
\end{center}
\end{subtable}

\begin{subtable}[t]{\textwidth}
\begin{center}
\caption{LibriSpeech}
\begin{tabular}{ccccccccccc}
\toprule
\multicolumn{11}{c}{Seen} \\
Speaker & 446 & 1246 & 2136 & 4813 & 4830 & 6836 & 7517 & 7800 & 8238 & 8123 \\
Gender & M & F & M & M & M & M & F & F & F & F \\
\midrule
\multicolumn{11}{c}{Unseen} \\
Speaker & 1320 & 2300 & 3570 & 3575 & 4970 & 4992 & 6829 & 7021 & 7729 & 8230 \\
Gender & M & M & F & F & F & F & F & M & M & M \\
\bottomrule
\end{tabular}
\end{center}
\end{subtable}
\end{small}
\end{center}
\end{table}

\section{Fictitious speakers}

\begin{figure*}[h]
  \begin{center}
    \includegraphics[width=0.98\textwidth]{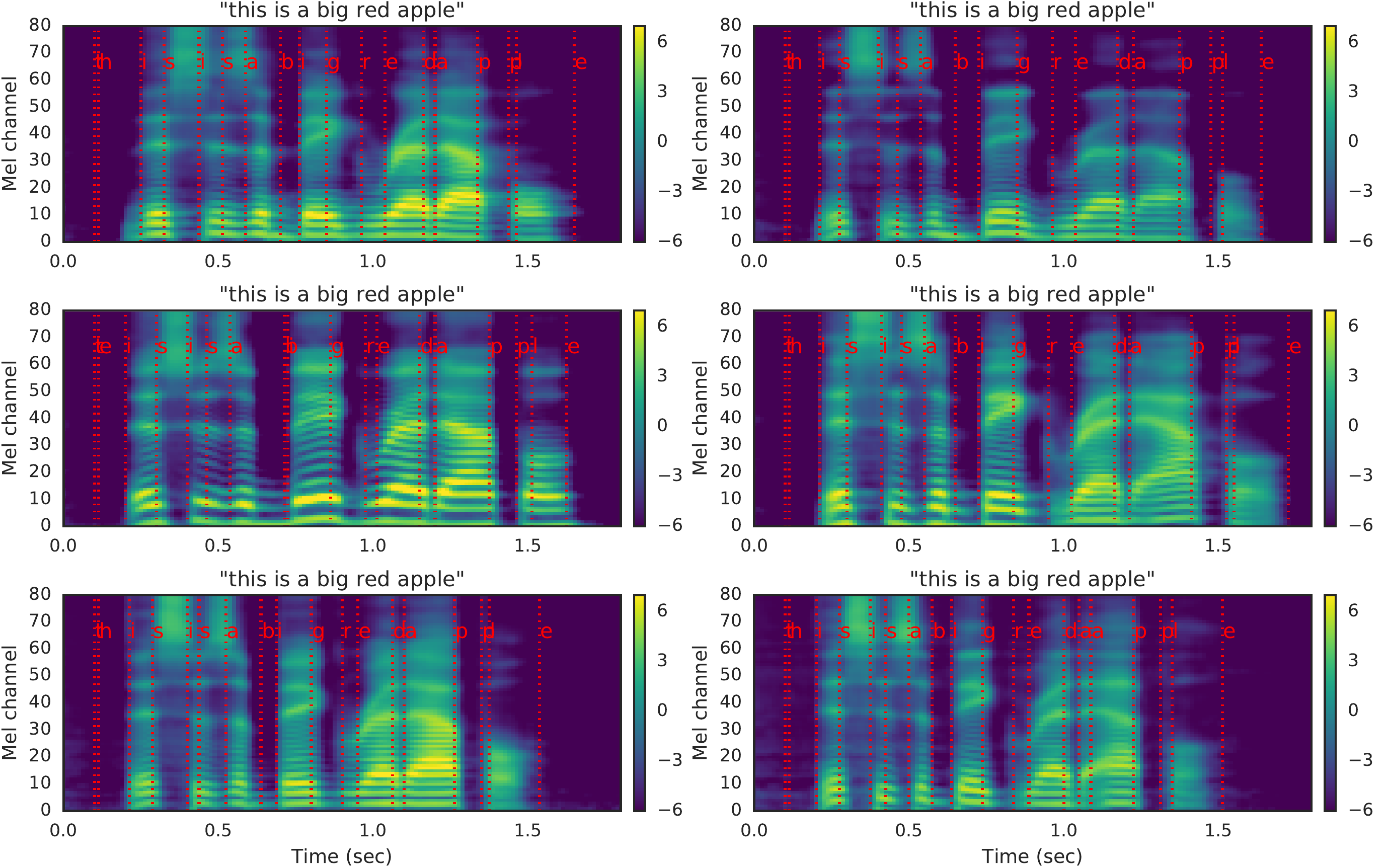}
    \caption{Example synthesis of a sentence conditioned on several random speaker embeddings sampled from the unit hypersphere.  All samples contain consistent phonetic content, but there is clear variation in fundamental frequency and speaking rate.
    Audio files corresponding to these utterances are included in the demo page (\url{https://google.github.io/tacotron/publications/speaker_adaptation}).
    }
  \label{fig.example_outputs_random_dvector}
  \end{center}
\end{figure*}

\newpage

\section{Speaker similarity MOS evaluation interface}
\label{appx:similarity_eval_template}

\begin{figure}[h!]
\centering
\includegraphics[width=\textwidth]{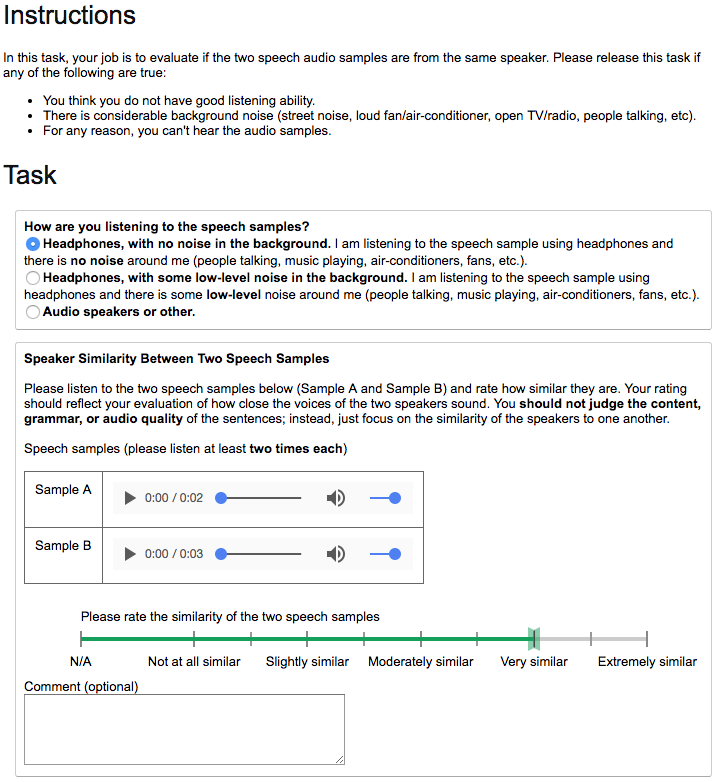}
\caption{Interface of MOS evaluation for speaker similarity.}
\label{fig:similarity_eval_template}
\end{figure}

\end{appendices} 
\end{document}